\documentclass[letterpaper, 10 pt, conference]{ieeeconf}

\IEEEoverridecommandlockouts                              

\usepackage{graphics} 
\usepackage{epsfig} 
\usepackage{mathptmx} 
\usepackage{times} 
\usepackage{amsmath} 
\usepackage{amssymb}  
\usepackage{lipsum}
\usepackage{caption}

\usepackage{graphicx}

\graphicspath{{images/}}

\title{\LARGE \bf
Can I Trust You? A User Study of Robot Mediation\\ of a Support Group
}

\author{Chris Birmingham$^{1}$, Zijian Hu$^{2}$, Kartik Mahajan$^{3}$, Eli Reber$^{4}$, and Maja  J Matari\'{c}$^{5}$
\thanks{*This work was supported by the National Science Foundation}
\thanks{$^{1}$Chris Birmingham is a PhD Student at the Interaction Lab, in the CS Department, University of Southern California, Los Angeles, CA, The United States
        {\tt\small cbirming@usc.edu}}%
\thanks{$^{2}$Zijian Hu is an Undergraduate Student with Interaction Lab, in the CS Department, University of Southern California, Los Angeles, CA, The United States
        {\tt\small zijianhu@usc.edu}}%
\thanks{$^{3}$Kartik Mahajan is an Undergraduate Student with Interaction Lab, in the CS Department, University of Southern California, Los Angeles, CA, The United States
        {\tt\small kmahajan@usc.edu}}%
\thanks{$^{4}$Eli Reber is an Undergraduate Student at 
        Pennsylvania State University, State College, PA, The United States
        {\tt\small Edr3@psu.edu}}%
\thanks{$^{5}$Maja J Matari\'{c} is the PI of the Interaction Lab, Professor of Computer Science, Neuroscience, and Pediatrics at the University of Southern California,
        {\tt\small mataric@usc.edu}}%
\thanks{© 2020 IEEE. Personal use of this material is permitted. Permission from IEEE must be obtained for all other uses, in any current or future media, including reprinting/republishing this material for advertising or promotional purposes, creating new collective works, for resale or redistribution to servers or lists, or reuse of any copyrighted component of this work in other works.}
}

\begin{document}

\maketitle
\thispagestyle{empty}
\pagestyle{empty}

\begin{abstract}
Socially assistive robots have the potential to improve group dynamics when interacting with groups of people in social settings. This work contributes to the understanding of those dynamics through a user study of trust dynamics in the novel context of a robot mediated support group. For this study, a novel framework for robot mediation of a support group was developed and validated. To evaluate interpersonal trust in the multi-party setting, a dyadic trust scale was implemented and found to be uni-factorial, validating it as an appropriate measure of general trust. The results of this study demonstrate a significant increase in average interpersonal trust after the group interaction session, and qualitative post-session interview data report that participants found the interaction helpful and successfully supported and learned from one other. The results of the study validate that a robot-mediated support group can improve trust among strangers and allow them to share and receive support for their academic stress.

\end{abstract}


\section{INTRODUCTION}

In a world where technology has increasingly involved users in virtual environments and isolated them in their local environments, social robots present an opportunity to engage people in situated pro-social ways. By their embodied nature, robots inherently invite people to come together and interact in physical spaces, while creating opportunities to assist and improve human-human interaction. Shaping and improving one-on-one interactions and group interaction dynamics are both specific goals of socially assistive robotics (SAR) \cite{feil2005defining, mataric2016socially}. 


A key research challenge of human-robot interaction (HRI) in general and SAR in particular is understanding interpersonal dynamics in group settings. In such settings, individuals interact with one another through complex verbal and nonverbal signals that are contextual and change over time. To successfully interact with and mediate group interactions, robots must be able to recognize human signals in real time, understand what they mean in a given context, and then choose  appropriate actions to achieve group goals that may involve improving cohesion, communication, engagement, or trust. Sensing and improving these properties of group dynamics is challenging because they involve the interaction of many, often subtle, multimodal signals \cite{mana2007multimodal}.

This work introduces HRI and SAR into the novel context of support group mediation. {\it Support groups} are meetings in which individuals with a common problem or challenge provide support to one another, typically with the help of a mediator \cite{jacobs2011group}. {\it Trust} is crucial for proper functioning of support groups, because it is only when participants feel that they are in a trustworthy place that they are  be willing to share and receive support \cite{johnson1972effects}. In most support groups, the level of trust changes over time as participants make disclosures and experience supportive responses from others. Group participants regularly evaluate and update their trust in one another as the session progresses \cite{corey2013groups}. Since trust in a support group setting can change relatively quickly and significantly, the context is both challenging and well suited for capturing data for training robots to learn the signals and dynamics associated with changes in trust. Although trust between participants in a support group typically grows over time, the skill of the mediator plays an important role in group success. Corey \& Corey \cite{corey2013groups} emphasize that mediators must have leadership skills such as genuineness, caring, openness, self-awareness, active listening, confronting, supporting, and modeling in order to lead a group effectively. They also point out how trust can be gained or lost by how the mediator copes with conflict or the initial expression of negative reactions. 

As an early step toward effective robot support group mediators, this exploratory work develops a novel framework for selecting robot mediator questions and disclosures, as well as a dyadic trust scale for measuring interpersonal trust. The framework is validated on a dataset collected in semi-structured interactions of 27 three-person robot-mediated support groups for academic stress. The study was conducted just prior to academic year-end final exams and involved genuinely supportive interactions among the participating students. The dyadic trust scale is found to be uni-factorial, validating it as an appropriate general measure of trust. The results show that robot-mediated interactions significantly increased dyadic trust between participants as well as between participants and the robot, and participants genuinely shared with one another under the guidance of the robot mediator.

\section{RELATED WORK}
This work aims to contribute to the understanding of group social dynamics in multi-party human-robot interaction.  Consequently, this section reviews relevant work in the broader multi-party human-robot interaction context, with special attention given to works in facilitating conversation and Wizard of Oz studies. Finally, this section provides a brief background on the definition and measures of trust.

\subsection{Multi-Party Human-Robot Interaction}
Within HRI, there is a growing body of work developing social robots for multi-party interactions. Social robots have been used in a wide variety of multi-party contexts, but the most popular are teaching, entertaining, serving, and mediation. In the teaching context, robots have been used for tutoring \cite{mussakhojayeva2016should}, for exhibit guiding \cite{salam2015engagement}, and for dispensing information \cite{bohus2014directions}. In the entertainment context, robots have been used to play games \cite{fraune2017teammates}, give presentations \cite{sugiyama2015estimating}, and participate in conversations \cite{vsabanovic2013paro}. In the service context, robots have been studied as waiters serving drinks in open settings \cite{vazquez2015parallel} and behind the bar \cite{kirchner2011nonverbal}. In the mediation context, robots have primarily been used for moderating game play \cite{short2017robot, short2017understanding, jung2015using} and mediating conversations. In this work the robot takes the role of mediator in a support group context. 

\subsubsection{Studies in Mediating Conversation}
This work follows prior work in which the robot acted as both a mediator and facilitator in order to improve the quality and balance of a conversation. Tahir et al. \cite{tahir2018user} focused on different methods for improving conversational quality using a robot mediator to deliver feedback about a conversation between two individuals. In that study, two individuals acted out a scripted conversation and then the robot delivers feedback created by a `sociofeedback system' which analyzed the interest, dominance, and agreement displayed in the conversation. Although the paper reported on the sociofeedback system training, the focus of the study was on the user's perceptions of the way the feedback was delivered via the robot, in which they found that the participants liked receiving feedback from the robot. Recent work has explored using a robot as a counselor in couples therapy to improve the quality of communication \cite{zuckerman2015empathy}, and promote positive communication \cite{utami2017robotic} and collaborative responses \cite{utami2019collaborative}.
Other work has focused on improving the balance and flow of conversations. One of the earlier efforts to explore conversational balance utilized a facilitation robot to obtain the conversational initiative and regulate imbalance \cite{matsuyama2015four}. The work of Short et al. \cite{short2016modeling} evaluated user perception of a robot mediator in a controlled study in which participants completed a group story telling task. In Ohshima et al. \cite{ohshima2017neut}, the authors tested robot behaviors for helping a group recover from an awkward silence. In their study, a robot led a conversation with three participants, taking actions and asking questions to encourage conversation.
The work presented here also utilizes the robot as a mediator for a conversation, in which the robot asks questions of each group member in order to elicit sharing. Unlike prior work, however, instead of attempting to maximize communication quality or conversational balance, the presented work examines how users relate to each other over the course of the support group interaction, particularly with respect to trust.

\subsubsection{Wizard of Oz Studies of Multi-Party SAR Mediation}
Because of the difficulty of understanding and interacting with natural language, most prior SAR work on conversational mediation has constrained the role of the robot and the action space in which the robot can participate. For instance, Hoffman et al. \cite{hoffman2015design} documented the design and evaluation of a nonverbal conversational companion that attempts to encourage empathy between the individuals having the conversation. Other work has constrained the interaction problem by ignoring the control challenges through a Wizard of Oz (WoZ) framework in which a hidden human controls the behavior of the robot. This paradigm can introduce confounds and bias into the interaction; however, it can be appropriate if reported correctly and used as part of an iterative approach to developing technology \cite{riek2012wizard}. For example, in the HRI work of Nigam \& Rick \cite{nigam2015social}, the interest was solely in collecting data for building classification models of the environment and therefor using WoZ was appropriate. In Vazquez et al. \cite{vazquez2017towards}, the authors were studying the human perception of the robot gaze and orientation behaviors, and so choose WoZ to control the timing and choice of actions the robot takes. Similarly, in the work presented here, WoZ is primarily employed to control the timing of programmed robot speech and actions, allowing the robot to interact smoothly in the conversation. This minimized WoZ control allows the collection of realistic interaction data for developing future autonomous control while also testing the baseline effects of the robot's mediation methodology.


\subsection{Trust in Support Group Mediation}
Trust is challenging to define and measure, yet it is crucial to the proper functioning of support groups. Trust can have different meanings depending on the context, however its definition in the literature has consolidated around three main factors: ability, benevolence, and integrity \cite{mayer1995integrative}. This work utilizes the definitions from Williams et al. \cite{williams2001whom}, where ability is defined as ``a set of skills that allow an individual to perform in some area," benevolence as ``the other-oriented desire to care for the protection of another," and integrity as ``the belief that another adheres to a set of principles that one finds acceptable". In support groups (and other therapeutic groups), Johnson \& Noonan \cite{johnson1972effects} identified benevolence and integrity as the more pertinent aspects of trust. Unfortunately, no validated measures have been developed for change in trust of group members over a single session support group interaction. Two relevant measures for individual trust are the Dyadic Trust Scale \cite{larzelere1980dyadic} and the Specific Interpersonal Trust Scale (SITS) \cite{johnson1982measurement}. The Dyadic Trust Scale is a uni-dimensional scale that focuses on measuring trust in close personal relationships. The SITS also measures trust in close personal relationships, but has been broken out to include factors of reliableness, emotional trust, and general trust. In support groups, the relevant factors are measured by the emotional trust and general trust subscales. Because it is unclear what measures would capture the short term change in trust, the participants in this work were given both measures and custom questions based on the established antecedents of trust: ability, benevolence, and integrity.

\section{METHOD}
\subsection{Participants}
A total of 81 university students who self-identified as \textit{stressed} participated in the study in groups of three; each group met once and the study produced a total of 27 recorded group sessions. For the purposes of the study, {\it stress} was defined as all forms of academic stress, including all concerns pertaining to class work and performance. All participants provided consent to participate in the study, which was approved under USC IRB UP-19-00084. The participant demographics were as follows: gender 48\% female, 49\%  male, 3\%  preferred not to specify; ethnicity; 74\% Asian, 11\% Hispanic/Latinx, 13\% Caucasian, 3\% African American; degree being pursued: 45\% undergraduate, 32\% master's, and 24\% PhD. 

\begin{figure}[ht]
    \centering
    \includegraphics[width=8cm]{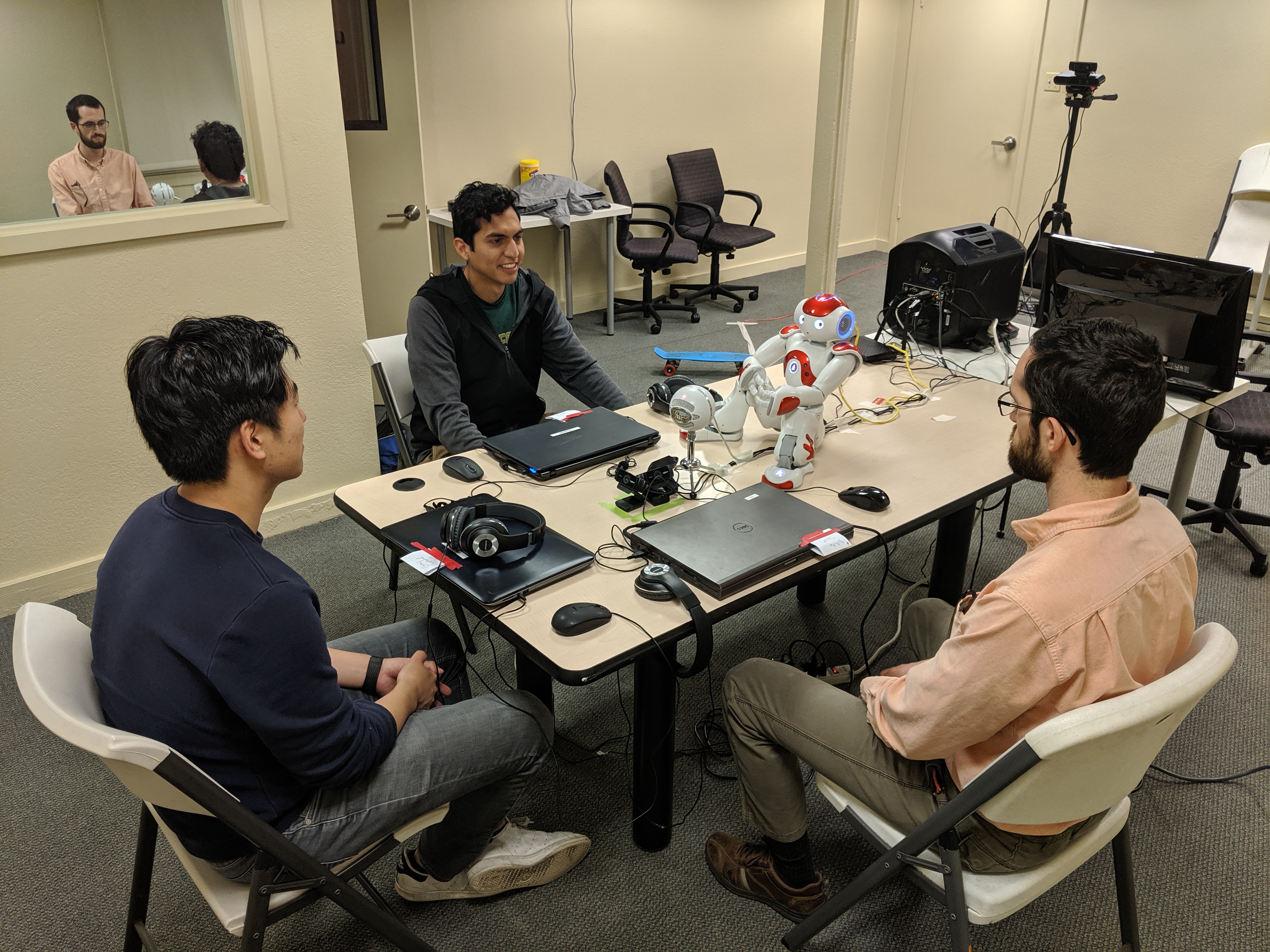}
    \caption{Volunteers demonstrating the study setup; participants could see one another and the robot but not each other's computer screens.}
    \label{fig:studyPhotoGroupSession}
\end{figure}

\subsection{Study Design}\label{setup}
\subsubsection{Physical Setup}
In each session, the three participants were seated around the end of a table with a seated Nao robot \cite{Nao2019} on it, as seen in Figure \ref{fig:studyPhotoGroupSession}. The Nao is a humanoid robot, 22.6" tall, with a total of 25 degrees of freedom. The robot was positioned as a member of the group, and served as the group moderator. Between the robot and the participants, a 360-degree microphone recorded audio data. At the base of the microphone, three HD webcams were arranged, one facing each of the participants; the webcams recorded participant body pose and facial expressions. Behind the robot, an RGB-D camera was mounted on a tripod and recorded the interaction of all four members of the group. The robot controller (Wizard) was seated behind a one-way mirror, out of the participants' sight. 
\begin{table*}[t!]
\begin{center}
\captionsetup{justification=centering}
\begin{tabular}{c c c}
 Sensitivity & Question & Disclosure\\
\hline
Low & What do you like about school? & When I feel stressed, I think my circuits might overload. \\ & & Does anyone else feel the same way?\\
\hline
Medium & What are some of the hardest parts of school for you? & Sometimes I worry I am inadequate for this school. \\ & & Does anyone else sometimes feel that too?\\
\hline
High & What will happen if you don't succeed in school? & Sometimes I worry about if I belong here. \\ & & Does anyone else feel the same way?\\
\hline
\end{tabular}
\caption{\small Example questions and disclosures spoken by the robot, indicating sensitivity. A total of 16 questions and 6 disclosures were available; an average of 12 questions and 3 disclosures were made by the robot in each session.}
\label{table:Questions}
\end{center}
\end{table*}

\subsubsection{Interaction Framework}
The robot's role as a mediator consisted of initiating and encouraging the process of sharing and supporting within the group, by asking questions that encouraged participants to share with one another. The Wizard controlled the robot's head direction (and therefore its gaze direction) to look at the speaking member of the group, and controlled the timing of the robot's speech (a question or disclosure) based on when the group finished the discussion of the previous topics/question. The robot's questions and disclosures were open-ended and specifically designed to encourage the participants to share with one another, in order to support and learn from one another. The content ranged from low sensitivity, such as, ``What do you like about school?" to high sensitivity, such as, ``Sometimes I worry about if I belong here, does anyone else feel the same way?" In the study, {\it sensitivity} was defined by how personal or invasive a question was or how uneasy a question might make a participant feel \cite{kaplan2015measuring}. 

During the interaction, the robot said questions and disclosures according to a simple algorithm. The questions and disclosures were grouped into low, medium, and high sensitivity as illustrated in Table \ref{table:Questions}, and the robot started with low sensitivity questions before moving onto medium and then high sensitivity. Before transitioning to the next highest level, the robot would make a disclosure. This pattern of questions and disclosures starting at low sensitivity and moving to high sensitivity allowed participants to become comfortable sharing and start trusting each other at each level of sensitivity.
The robot alternated between questions and disclosures and balanced who in the group received each question, to balance the number of questions first posed to each participant. The robot maintained a neutral affect, with no facial expression and neutral tonal affect.

\subsection{Procedure}
After participants consented to take part in the study, they were seated as shown in Figure \ref{fig:studyPhotoGroupSession}, answered a few demographic questions, and completed the pre-study trust survey consisting of 30 Likert scale questions (described in Section \ref{prepost}). Participant's familiarity with each other was not measured.  The robot then began the group interaction by explaining that the purpose of the session was for them to talk about their academic stress and help one another. The robot then asked the participants to introduce themselves.  After the introductions, the Wizard took control of the robot for the remaining ~20 minutes of the session and operated chose questions and disclosures according to the framework described above. At the end of the session, the robot asked the group to conclude the group session by sharing what they felt they had learned. 

After the group interaction was complete, the participants completed the trust survey again. Participants were then invited to take part in an open-ended group interview, in which they had the opportunity to provide feedback about their experience. Finally, the participants completed a custom survey assessing their baseline trust ("Would you say that most people can be trusted or that you can never be too careful with people?"), the Negative attitudes towards Robots Survey \cite{nomura2004psychology}, the Big Five (short) Inventory \cite{rammstedt2007measuring}, and the Empathy Inventory \cite{davis1980multidimensional}.

\subsection{Trust Surveys} \label{prepost}
A battery of trust surveys was administered to evaluate the pre- and post-study levels of participant trust. Three validated surveys were used: 1) the Dyadic Trust Scale \cite{larzelere1980dyadic}, and the Specific Interpersonal Trust Scale 2) Overall Subscale and 3) Emotional Subscale \cite{johnson1982measurement}. Additionally, a customized study-specific scale was administered, consisting of six questions based on the antecedents of trust: benevolence, integrity, and ability (see \cite{colquitt2007trust} for a meta-review of their significance). The complete combined battery of surveys consisted of 30 Likert scale questions; each participant completed three copies of the identical battery of surveys before and after the group session: one survey was about the robot and the other two were about the other two study participants.

\section{RESULTS}

\begin{table*}[t!]
\begin{center}
\captionsetup{justification=centering}
\begin{tabular}{c c c c c c c c c}
 & Agreeableness & Conscientiousness & Extroversion & Neuroticism & Openness & Emp Overall & NARS Overall & Trust Baseline\\
\hline
Robot Before & \textbf{0.25} & -0.13 & 0.11 & -0.18 & 0.09 & 0.01 & \textbf{-0.37} & \textbf{0.27}\\
\hline
Robot Change & -0.11 & \textbf{0.34} & -0.02 & 0.17 & 0.01 & -0.10 & -0.12 & 0.03\\
\hline
Group Before & 0.08  & -0.07 & -0.04 & 0.20 & 0.10 & 0.07 & -0.20 & -0.06\\
\hline
Group Change & 0.21 & -0.07 & -0.01 & -0.13 & -0.05 & 0.18 & 0.06 & -0.04\\
\hline
\end{tabular}
\caption{\small Correlation table for subscales and trust, p\textless0.05 are bolded}
\label{table:Correlations}
\end{center}
\end{table*}

This section presents the quantitative and qualitative results of the study. These results are based on the scores of 71 participants; the scores of 10 participants who did not complete the surveys or filled them in without coding the reverse questions correctly were removed, resulting in n=71. The survey scores from each battery of surveys were combined to produce three numeric scores on a scale -3 to 3 indicating the level of trust the participant felt towards each of the other participants, and the robot.

\subsection{Overall Trust}

\begin{figure}[ht]
    \centering
    \includegraphics[width=8cm]{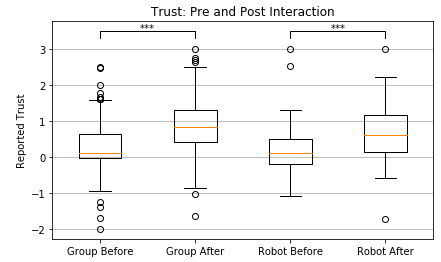}
    \caption{Box plot of participants' trust pre- and post-interaction, relative to the other group members and the robot.  Medians are shown in orange. }
    \label{fig:trust_box}
\end{figure}

To determine significance, a t-test for comparing paired samples on the values of trust was conducted before and after the support group session. The effect size was calculated by
\[ r = \frac{Z}{\sqrt{N}}\]
 where Z was the standardized test statistic from a Wilcoxon Signed Rank Test and N was the size of the corresponding population.  For this analysis it is assumed that each participants' rating of trust in the other two participants was independent, doubling the population (n=142) as compared to the robot (n=71). As shown in Figure \ref{fig:trust_box}, there was a significant increase in trust participants felt in one another and in the robot; the effect size of the increase in trust was large for both groups. 

\begin{table}[ht]
\label{significance_table}
\begin{center}
\captionsetup{justification=centering}
\begin{tabular}{c c c c c}
\multicolumn{1}{c}{} & \multicolumn{1}{c}{T-value} & \multicolumn{1}{c}{P-value} & \multicolumn{1}{c}{Effect Size} & \multicolumn{1}{c}{95$\%$ CI for $\Delta$ Mean}  \\
\hline
Group t-(141) & -9.46 & \textbf{\textless  0.001} & 0.66 & (0.24, 0.73)\\ 
\hline
Robot t-(70) & -6.95 & \textbf{\textless  0.001} & 0.68 & (0.43, 0.77)\\
\hline
\end{tabular}
\caption{\small Test statistics for the overall change in trust given to each corresponding entity by the session participants}
\end{center}
\end{table}

\begin{figure}[h!]
    \begin{center}
    \includegraphics{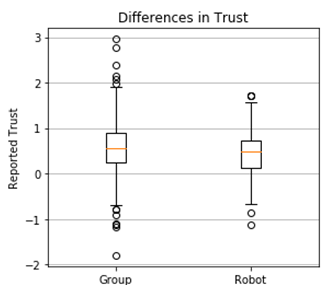}
    \caption{Box plot of the change in trust participants felt in group members and in the robot, illustrating the effect the group interaction had on trust. Medians are shown in orange.}
    \label{fig:Differences}
    \end{center}
\end{figure}

Figure \ref{fig:trust_box} shows that the distributions of the group and robot trust contained similar changes in the mean (increases of approximately 0.65 and 0.50, respectively), but different changes in standard deviation (increases of 0.01 and 0.13, respectively).  This highlights the growing variability of the participants' trust for the robot, as compared to their responses to the other participants. It can also be seen from Figure \ref{fig:Differences} that the \textit{differences} in the robot trust (\textit{SD} = 0.58) were less variable than those of group trust (\textit{SD} = 0.75).  This was most likely due to the larger number of outliers in the group differences distribution than the robot differences distribution (13 and 5, respectively). Aside from these, there are no other statistically significant difference between the two distributions (p = 0.14).

The trust levels for the robot also provide support for its use in the support group context. The large effect size and the significant increase in trust point towards the robot's potential as an effective mediator in support group conversations.

\subsection{Demographics}
After using a Bonferroni correction for multiple comparisons, none of the tested demographics (age, gender, ethnicity, and degree sought) had a statistically significant impact on the trust in the robot or other participants in the session. As can be seen in Table \ref{table:Correlations}, there were significant correlations between trust in the robot before and Agreeableness and Trust Baseline (.25) as well as a strong negative correlation with NARS Overall (-.36). Interestingly, the only correlation with the Robot Change was Conscientiousness, possibly due to a sense of duty to rate the robot higher after the group interaction. There were no significant correlations in any of the surveys with Group Before or Group Change.

\subsection{Factor Analysis}

A factor analysis was performed to understand the hidden latent variables affecting the results for n =71 participants.  The Root Mean Squared Error of Approximation and Root Mean Square of Residuals were used as an adequacy test to determine if a number of factors was sufficient \cite{FactorAnalysis}. One dominant factor sufficient to meet the adequacy test standards (RMSEA=(0.04, 0.06), RMSR=(0.04, 0.06)) was found. Removing the participants who did not take the survey seriously resulted in higher inter-correlations among the questions, thus allowing more questions to be loaded into the factor.  After cleaning the results, the survey data showed a total Cronbach's alpha of over 0.9.  The robot and group trust subscales also showed Crobach's alphas of over 0.9, thus validating the internal consistency of the participants' responses.

\begin{figure}[h!]
    \begin{center}
    \includegraphics[width=8.7cm]{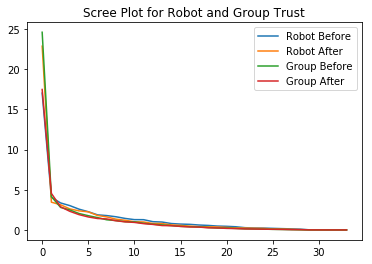}
    \caption{Scree plot of principal components in the composite trust survey for group members and the robot, before and after the interaction. All plots show a strong elbow at the second factor, indicating one factor explains most of the variance.}
    \label{fig:Scree}
    \end{center}
\end{figure}

The questions that were loaded into the factor held an overarching theme of general trust. The single trust factor became more dominant after the session for the robot trust questions (eigenvalue went from 17.03 to 22.87), but became less dominant for the group trust questions (eigenvalue went from 24.60 to 17.48). This suggests that the participant conceptions of trust in the robot became more monolithic (meaning uniform and indivisible) after the interaction, while conceptions of trust in the other participants became less monolithic. All but two of the survey questions were loaded into the one factor.  In looking at the unloaded questions, it can be seen that they also measure general trust, but include elements of the ability aspect of trust that may have deterred participants from having similar answers to other questions. For example, the question ``I would trust the robot to take me to the airport" had very low correlations with the other questions, even though it is asking about the participants' trust in the robot because it seems infeasible that the small robot could drive. After performing the factor analysis, it was decided that two questions that were poorly correlated with the rest of the survey (less than 0.3) would be thrown out. The single factor solution suggests that participants had a vague, monolithic notion of trust in one another that they used to answer the survey. This may be because even after the group interaction, they had only known one another for a short time, and had yet to solidify more distinct facets of trust in one another and the robot.

\subsection{Qualitative Results}
At the conclusion of the support group setting the participants were asked by the robot, ``What is something each of you learned today?" Many participants expressed sentiments that they were ``not alone in feeling stressed" and ``everyone is in the same boat." Several expressed that they felt they had ``learned new tips and strategies for dealing with stress." Not all participants described feeling that way. Several chose to focus more on the robot, for example saying``I am much more comfortable talking to the robot than I thought I would be" and ``I am shocked by machine and human interaction, the robot can talk and kind of understand feelings." 

In their optional feedback, many participants expressed that they enjoyed the group interaction. Contrary to expectations, almost all participants expressed that stress was not a sensitive topic, and that they ``talk about academic stress all the time" with family and friends. When asked how the session differed from everyday conversations about school stress, some participants focused on how the discussion with the robot was mechanical and not a free flowing discussion, while others focused on how they felt talking with a robot and strangers allowed them to ``say things I could not share in other situations." Although most participants expressed that it was easy to talk about academic stress, even with strangers and a robot, almost all participants said they felt they grew closer with one another through the group interaction, and that trust grew as they shared with one another. Supported by the survey data, this validates the hope that robot-mediated support group interaction increases participant trust and helps go alleviate academic stress.

In the group interview, participants also offered feedback on the limitations of the interaction and suggestions for improvement. A commonly-discussed limitation many was the ``lack of humanity" of the Nao robot as a mediator. This sentiment was explained by several participants as being related to the Nao robot's simple, inexpressive face. Although the robot employed gestures such as shrugging and head scratching, participants felt that it could not display empathy and that the pauses and gestures were `awkward'. The robot turned its head to look at the participant who was speaking; one participant felt ``it was always watching me" while others described it as `lifeless'. Another limitation participants identified in the design was the robot sound. Participants felt that the noise of the robot's motors interrupted the conversation flow and reminded them that the mediator was a robot. Non-native English speakers also had trouble understanding the robotic voice and often asked for the question and disclosures to be repeated. When discussing suggestions for improvement, two common themes emerged. The first was the participants' wish for the robot to do more than ask questions and make disclosures: they suggested that the robot should acknowledge or follow up on what was said before moving on with the conversation. The second theme was the desire for the robot to either share more of its own back story and be able to answer questions about itself, or to state at the beginning that its purpose was solely to mediate and that it would not answer questions.

\section{CONCLUSION}
This work reported on a user study evaluating trust in a socially assistive robot-mediated academic support group for stressed university students. A group mediation framework was developed and validated. To measure trust, a dyadic trust scale was implemented and found to be uni-factorial, validating it as an appropriate general measure of trust. The support group interaction had a strong effect on participants' trust in one another and on their trust in the robot. Participants were willing to learn from and share with one another under the guidance of the robot mediator. This work validated that the robot did function as an effective mediator of the support group interaction, opening the door to future work with robot support group mediators. 

Future work will utilize the collected data to develop models of the multi-party human-robot interaction dynamics. It will also use the self-annotations of trust to develop predictive models of how trust changes in order to inform autonomous robot control. Based on the participant feedback, ongoing work will explore using more human-like voices to improve participant comprehension of the robot's speech. The collected data will be used to develop improved models of turn-taking that will allow the robot to choose when to speak autonomously and allow it to provide acknowledgement after a participant finishes speaking in their turn. The presented work aims to inform the development of effective robot mediators that can positively influence and improve group dynamics in compelling settings such as support groups.

\addtolength{\textheight}{-.2cm}


\section*{ACKNOWLEDGMENT}

This work was supported by the National Science Foundation Expeditions in Computing IIS-1139148. The authors thank Dr. Matt Rueben for his help with data analysis.


\newpage
\bibliographystyle{IEEEtran}
\bibliography{IEEEabrv,IEEEexample,Understanding_Trust,DataCollection}

\begin{thebibliography}{10}
\providecommand{\url}[1]{#1}
\csname url@rmstyle\endcsname
\providecommand{\newblock}{\relax}
\providecommand{\bibinfo}[2]{#2}
\providecommand\BIBentrySTDinterwordspacing{\spaceskip=0pt\relax}
\providecommand\BIBentryALTinterwordstretchfactor{4}
\providecommand\BIBentryALTinterwordspacing{\spaceskip=\fontdimen2\font plus
\BIBentryALTinterwordstretchfactor\fontdimen3\font minus
  \fontdimen4\font\relax}
\providecommand\BIBforeignlanguage[2]{{%
\expandafter\ifx\csname l@#1\endcsname\relax
\typeout{** WARNING: IEEEtran.bst: No hyphenation pattern has been}%
\typeout{** loaded for the language `#1'. Using the pattern for}%
\typeout{** the default language instead.}%
\else
\language=\csname l@#1\endcsname
\fi
#2}}

\bibitem{feil2005defining}
D.~Feil-Seifer and M.~J. Mataric, ``Defining socially assistive robotics,'' in
  \emph{9th International Conference on Rehabilitation Robotics, 2005. ICORR
  2005.}\hskip 1em plus 0.5em minus 0.4em\relax IEEE, 2005, pp. 465--468.

\bibitem{mataric2016socially}
M.~J. Matari{\'c} and B.~Scassellati, ``Socially assistive robotics,'' in
  \emph{Springer Handbook of Robotics}.\hskip 1em plus 0.5em minus 0.4em\relax
  Springer, 2016, pp. 1973--1994.

\bibitem{mana2007multimodal}
N.~Mana, B.~Lepri, P.~Chippendale, A.~Cappelletti, F.~Pianesi, P.~Svaizer, and
  M.~Zancanaro, ``Multimodal corpus of multi-party meetings for automatic
  social behavior analysis and personality traits detection,'' in
  \emph{Proceedings of the 2007 workshop on Tagging, mining and retrieval of
  human related activity information}.\hskip 1em plus 0.5em minus 0.4em\relax
  ACM, 2007, pp. 9--14.

\bibitem{jacobs2011group}
E.~E. Jacobs, R.~L. Masson, R.~L. Harvill, and C.~J. Schimmel, \emph{Group
  counseling: Strategies and skills}.\hskip 1em plus 0.5em minus 0.4em\relax
  Cengage learning, 2011.

\bibitem{johnson1972effects}
D.~W. Johnson and M.~P. Noonan, ``Effects of acceptance and reciprocation of
  self-disclosures on the development of trust.'' \emph{Journal of Counseling
  Psychology}, vol.~19, no.~5, p. 411, 1972.

\bibitem{corey2013groups}
M.~S. Corey, G.~Corey, and C.~Corey, \emph{Groups: Process and practice}.\hskip
  1em plus 0.5em minus 0.4em\relax Cengage Learning, 2013.

\bibitem{mussakhojayeva2016should}
S.~Mussakhojayeva, M.~Zhanbyrtayev, Y.~Agzhanov, and A.~Sandygulova, ``Who
  should robots adapt to within a multi-party interaction in a public space?''
  in \emph{The Eleventh ACM/IEEE International Conference on Human Robot
  Interaction}.\hskip 1em plus 0.5em minus 0.4em\relax IEEE Press, 2016, pp.
  483--484.

\bibitem{salam2015engagement}
H.~Salam and M.~Chetouani, ``Engagement detection based on mutli-party cues for
  human robot interaction,'' in \emph{2015 International Conference on
  Affective Computing and Intelligent Interaction (ACII)}.\hskip 1em plus 0.5em
  minus 0.4em\relax IEEE, 2015, pp. 341--347.

\bibitem{bohus2014directions}
D.~Bohus, C.~W. Saw, and E.~Horvitz, ``Directions robot: in-the-wild
  experiences and lessons learned,'' in \emph{Proceedings of the 2014
  international conference on Autonomous agents and multi-agent systems}.\hskip
  1em plus 0.5em minus 0.4em\relax International Foundation for Autonomous
  Agents and Multiagent Systems, 2014, pp. 637--644.

\bibitem{fraune2017teammates}
M.~R. Fraune, S.~{\v{S}}abanovi{\'c}, and E.~R. Smith, ``Teammates first:
  Favoring ingroup robots over outgroup humans,'' in \emph{2017 26th IEEE
  International Symposium on Robot and Human Interactive Communication
  (RO-MAN)}.\hskip 1em plus 0.5em minus 0.4em\relax IEEE, 2017, pp. 1432--1437.

\bibitem{sugiyama2015estimating}
T.~Sugiyama, K.~Funakoshi, M.~Nakano, and K.~Komatani, ``Estimating response
  obligation in multi-party human-robot dialogues,'' in \emph{2015 IEEE-RAS
  15th International Conference on Humanoid Robots (Humanoids)}.\hskip 1em plus
  0.5em minus 0.4em\relax IEEE, 2015, pp. 166--172.

\bibitem{vsabanovic2013paro}
S.~{\v{S}}abanovi{\'c}, C.~C. Bennett, W.-L. Chang, and L.~Huber, ``Paro robot
  affects diverse interaction modalities in group sensory therapy for older
  adults with dementia,'' in \emph{2013 IEEE 13th International Conference on
  Rehabilitation Robotics (ICORR)}.\hskip 1em plus 0.5em minus 0.4em\relax
  IEEE, 2013, pp. 1--6.

\bibitem{vazquez2015parallel}
M.~V{\'a}zquez, A.~Steinfeld, and S.~E. Hudson, ``Parallel detection of
  conversational groups of free-standing people and tracking of their
  lower-body orientation,'' in \emph{2015 IEEE/RSJ International Conference on
  Intelligent Robots and Systems (IROS)}.\hskip 1em plus 0.5em minus
  0.4em\relax IEEE, 2015, pp. 3010--3017.

\bibitem{kirchner2011nonverbal}
N.~Kirchner, A.~Alempijevic, and G.~Dissanayake, ``Nonverbal robot-group
  interaction using an imitated gaze cue,'' in \emph{Proceedings of the 6th
  international conference on Human-robot interaction}.\hskip 1em plus 0.5em
  minus 0.4em\relax ACM, 2011, pp. 497--504.

\bibitem{short2017robot}
E.~Short and M.~J. Mataric, ``Robot moderation of a collaborative game: Towards
  socially assistive robotics in group interactions,'' in \emph{2017 26th IEEE
  International Symposium on Robot and Human Interactive Communication
  (RO-MAN)}.\hskip 1em plus 0.5em minus 0.4em\relax IEEE, 2017, pp. 385--390.

\bibitem{short2017understanding}
E.~S. Short, K.~Swift-Spong, H.~Shim, K.~M. Wisniewski, D.~K. Zak, S.~Wu,
  E.~Zelinski, and M.~J. Matari{\'c}, ``Understanding social interactions with
  socially assistive robotics in intergenerational family groups,'' in
  \emph{2017 26th IEEE International Symposium on Robot and Human Interactive
  Communication (RO-MAN)}.\hskip 1em plus 0.5em minus 0.4em\relax IEEE, 2017,
  pp. 236--241.

\bibitem{jung2015using}
M.~F. Jung, N.~Martelaro, and P.~J. Hinds, ``Using robots to moderate team
  conflict: the case of repairing violations,'' in \emph{Proceedings of the
  Tenth Annual ACM/IEEE International Conference on Human-Robot
  Interaction}.\hskip 1em plus 0.5em minus 0.4em\relax ACM, 2015, pp. 229--236.

\bibitem{tahir2018user}
Y.~Tahir, J.~Dauwels, D.~Thalmann, and N.~M. Thalmann, ``A user study of a
  humanoid robot as a social mediator for two-person conversations,''
  \emph{International Journal of Social Robotics}, pp. 1--14, 2018.

\bibitem{zuckerman2015empathy}
O.~Zuckerman and G.~Hoffman, ``Empathy objects: Robotic devices as conversation
  companions,'' in \emph{Proceedings of the ninth international conference on
  tangible, embedded, and embodied interaction}.\hskip 1em plus 0.5em minus
  0.4em\relax ACM, 2015, pp. 593--598.

\bibitem{utami2017robotic}
D.~Utami, T.~W. Bickmore, and L.~J. Kruger, ``A robotic couples counselor for
  promoting positive communication,'' in \emph{2017 26th IEEE International
  Symposium on Robot and Human Interactive Communication (RO-MAN)}.\hskip 1em
  plus 0.5em minus 0.4em\relax IEEE, 2017, pp. 248--255.

\bibitem{utami2019collaborative}
D.~Utami and T.~Bickmore, ``Collaborative user responses in multiparty
  interaction with a couples counselor robot,'' in \emph{2019 14th ACM/IEEE
  International Conference on Human-Robot Interaction (HRI)}.\hskip 1em plus
  0.5em minus 0.4em\relax IEEE, 2019, pp. 294--303.

\bibitem{matsuyama2015four}
Y.~Matsuyama, I.~Akiba, S.~Fujie, and T.~Kobayashi, ``Four-participant group
  conversation: A facilitation robot controlling engagement density as the
  fourth participant,'' \emph{Computer Speech \& Language}, vol.~33, no.~1, pp.
  1--24, 2015.

\bibitem{short2016modeling}
E.~Short, K.~Sittig-Boyd, and M.~J. Mataric, ``Modeling moderation for
  multi-party socially assistive robotics,'' in \emph{IEEE Int. Symp. Robot
  Hum. Interact. Commun.(RO-MAN 2016). New York, NY: IEEE}, 2016.

\bibitem{ohshima2017neut}
N.~Ohshima, R.~Fujimori, H.~Tokunaga, H.~Kaneko, and N.~Mukawa, ``Neut: Design
  and evaluation of speaker designation behaviors for communication support
  robot to encourage conversations,'' in \emph{2017 26th IEEE International
  Symposium on Robot and Human Interactive Communication (RO-MAN)}.\hskip 1em
  plus 0.5em minus 0.4em\relax IEEE, 2017, pp. 1387--1393.

\bibitem{hoffman2015design}
G.~Hoffman, O.~Zuckerman, G.~Hirschberger, M.~Luria, and T.~Shani~Sherman,
  ``Design and evaluation of a peripheral robotic conversation companion,'' in
  \emph{Proceedings of the Tenth Annual ACM/IEEE International Conference on
  Human-Robot Interaction}.\hskip 1em plus 0.5em minus 0.4em\relax ACM, 2015,
  pp. 3--10.

\bibitem{riek2012wizard}
L.~D. Riek, ``Wizard of oz studies in hri: a systematic review and new
  reporting guidelines,'' \emph{Journal of Human-Robot Interaction}, vol.~1,
  no.~1, pp. 119--136, 2012.

\bibitem{nigam2015social}
A.~Nigam and L.~D. Riek, ``Social context perception for mobile robots,'' in
  \emph{2015 IEEE/RSJ international conference on intelligent robots and
  systems (IROS)}.\hskip 1em plus 0.5em minus 0.4em\relax IEEE, 2015, pp.
  3621--3627.

\bibitem{vazquez2017towards}
M.~V{\'a}zquez, E.~J. Carter, B.~McDorman, J.~Forlizzi, A.~Steinfeld, and S.~E.
  Hudson, ``Towards robot autonomy in group conversations: Understanding the
  effects of body orientation and gaze,'' in \emph{Proceedings of the 2017
  ACM/IEEE International Conference on Human-Robot Interaction}.\hskip 1em plus
  0.5em minus 0.4em\relax ACM, 2017, pp. 42--52.

\bibitem{mayer1995integrative}
R.~C. Mayer, J.~H. Davis, and F.~D. Schoorman, ``An integrative model of
  organizational trust,'' \emph{Academy of management review}, vol.~20, no.~3,
  pp. 709--734, 1995.

\bibitem{williams2001whom}
M.~Williams, ``In whom we trust: Group membership as an affective context for
  trust development,'' \emph{Academy of management review}, vol.~26, no.~3, pp.
  377--396, 2001.

\bibitem{larzelere1980dyadic}
R.~E. Larzelere and T.~L. Huston, ``The dyadic trust scale: Toward
  understanding interpersonal trust in close relationships,'' \emph{Journal of
  Marriage and the Family}, pp. 595--604, 1980.

\bibitem{johnson1982measurement}
C.~Johnson-George and W.~C. Swap, ``Measurement of specific interpersonal
  trust: Construction and validation of a scale to assess trust in a specific
  other.'' \emph{Journal of personality and Social Psychology}, vol.~43, no.~6,
  p. 1306, 1982.

\bibitem{Nao2019}
\BIBentryALTinterwordspacing
``Nao the humanoid robot: Softbank robotics emea,'' 2019. [Online]. Available:
  \url{https://www.softbankrobotics.com/emea/en/nao}
\BIBentrySTDinterwordspacing

\bibitem{kaplan2015measuring}
R.~Kaplan and E.~Yu, ``Measuring question sensitivity,'' \emph{American
  Association for Public Opinion Research}, pp. 4107--4121, 2015.

\bibitem{nomura2004psychology}
T.~Nomura, T.~Kanda, T.~Suzuki, and K.~Kato, ``Psychology in human-robot
  communication: An attempt through investigation of negative attitudes and
  anxiety toward robots,'' in \emph{RO-MAN 2004. 13th IEEE International
  Workshop on Robot and Human Interactive Communication (IEEE Catalog No.
  04TH8759)}.\hskip 1em plus 0.5em minus 0.4em\relax IEEE, 2004, pp. 35--40.

\bibitem{rammstedt2007measuring}
B.~Rammstedt and O.~P. John, ``Measuring personality in one minute or less: A
  10-item short version of the big five inventory in english and german,''
  \emph{Journal of research in Personality}, vol.~41, no.~1, pp. 203--212,
  2007.

\bibitem{davis1980multidimensional}
M.~H. Davis \emph{et~al.}, ``A multidimensional approach to individual
  differences in empathy,'' 1980.

\bibitem{colquitt2007trust}
J.~A. Colquitt, B.~A. Scott, and J.~A. LePine, ``Trust, trustworthiness, and
  trust propensity: A meta-analytic test of their unique relationships with
  risk taking and job performance.'' \emph{Journal of applied psychology},
  vol.~92, no.~4, p. 909, 2007.

\bibitem{FactorAnalysis}
\BIBentryALTinterwordspacing
K.~J. Preacher, G.~Zhang, C.~Kim, and G.~Mels, ``Choosing the optimal number of
  factors in exploratory factor analysis: A model selection perspective,''
  \emph{Multivariate Behavioral Research}, vol.~48, no.~1, pp. 28--56, 2013,
  pMID: 26789208. [Online]. Available:
  \url{https://doi.org/10.1080/00273171.2012.710386}
\BIBentrySTDinterwordspacing

\end{thebibliography}

\end{document}